\documentclass[conference]{IEEEtran}
\IEEEoverridecommandlockouts
\usepackage{cite}
\usepackage[utf8]{inputenc}
\usepackage[colorlinks=true,linkcolor=black,urlcolor=blue,citecolor=black]{hyperref}
\usepackage{amsmath,amssymb,amsfonts}
\usepackage{algorithmic}
\usepackage{graphicx}
\usepackage{textcomp}
\usepackage{booktabs,multirow}

\usepackage[table]{xcolor}
\def\BibTeX{{\rm B\kern-.05em{\sc i\kern-.025em b}\kern-.08em
    T\kern-.1667em\lower.7ex\hbox{E}\kern-.125emX}}
\usepackage{todonotes}

\begin{document}

\title{Guiding the Creation of Deep Learning-based Object Detectors*\\
%Simplifying the use of YOLO thanks to Jupyter Notebooks
\thanks{*This work was partially supported by Ministerio de Econom\'ia y 
Competitividad [MTM2017-88804-P]. We also gratefully acknowledge the support of NVIDIA Corporation with the donation of the Titan Xp GPU used for this research.}
}

\author{\IEEEauthorblockN{Ángela Casado}
\IEEEauthorblockA{\textit{Departamento de Matemáticas y Computación} \\
\textit{Universidad de La Rioja}\\
Logroño, España \\
acg181293@hotmail.com}
\and
\IEEEauthorblockN{Jónathan Heras}
\IEEEauthorblockA{\textit{Departamento de Matemáticas y Computación} \\
\textit{Universidad de La Rioja}\\
Logroño, España \\
jonathan.heras@unirioja.es}
}

\maketitle

\begin{abstract}

Object detection is a computer vision field that has applications in several contexts ranging from biomedicine and agriculture to security. In the last years, several deep learning techniques have greatly improved object detection models. Among those techniques, we can highlight the YOLO approach, that allows the construction of accurate models that can be employed in real-time applications. However, as most deep learning techniques, YOLO has a steep learning curve and creating models using this approach might be challenging for non-expert users. In this work, we tackle this problem by constructing a suite of Jupyter notebooks that democratizes the construction of object detection models using YOLO. The suitability of our approach has been proven with a dataset of stomata images where we have achieved a mAP of 90.91$\%$.

\end{abstract}


\section{Introduction}

Object detection is a computer vision area that focuses on identifying the 
position of multiple objects in an image. Traditionally, object detection 
methods have been based on two main techniques known as sliding windows 
and image pyramids; and, they have been successfully applied to solve problems 
such as face detection~\cite{ViolaJones01} or pedestrian 
detection~\cite{DalalTrigs05}. However, 
those approaches are slow, lack the notion of aspect ratio and are error 
prone~\cite{Rosebrock17}. These problems have been recently overcome using deep learning techniques.

Deep learning has impacted almost every area of computer vision, and object detection is no exception. 
Intuitively, in deep learning-based object 
detectors, we input an image to a network and obtain, as output, the bounding 
boxes (that is, the minimum rectangle containing the objects) and the class labels. 
Deep learning-based object detectors can be split into two groups: one-stage and two-stage
object detectors. The former divide the image into regions, that are 
passed into a convolutional neural network, and then the prediction is obtained ---
these detectors include techniques such as SSD~\cite{SSD} or YOLO~\cite{yolov3}. 
The two-stage object detectors employ region proposal methods to obtain interesting 
regions in the image, that are later processed to obtain the prediction --- these 
methods include the R-CNN family of object 
detectors~\cite{Girshick14,Girshick15,Ren15}. 

Usually, one-stage object detectors are faster but less precise than two-stage 
object detectors; however, the one-stage object detector YOLO 
has recently achieved a similar accuracy to the one obtained by two-stage object 
detectors, but keeping fast processing~\cite{yolov3}. Due to this fact, the 
YOLO object detector has been applied in problems that require real time processing 
but also need a high accuracy, like real time detection of lung 
modules~\cite{Ramachandran18} or the detection of small objects in satellite 
imagery~\cite{Etten18}. 

The YOLO object detector is provided as part of the Darknet framework~\cite{darknet}; but, as almost every deep learning tool, 
it has a steep learning curve and adapting it to work in a particular problem might be a challenge for several users --- apart from understanding
the underlying algorithm of YOLO (a step that might not be necessary to create an object detection model with YOLO), using YOLO might be a challenge since it requires, among others, the installation of
several libraries, the modification of several files and the usage of a concrete file structure. 
Therefore, even if this technique might be helpful for several fields (ranging from biomedicine and agriculture to security), users from those fields are not able to take
advantage of it. To solve this problem, we have developed an assistant, in the form of a suite of Jupyter notebooks, that guides non-expert users through all the
steps that are necessary in an object detection project using YOLO.

As we explain in Section~\ref{sec:workflow}, there are several steps that are required to create a deep learning-based object detector; but, thanks to our assistant, see Section~\ref{sec:jupyter}, they are reduced to fixing a few parameters, and the rest of the process is conducted automatically by our tool. 
To prove the suitability of our approach, we have employed the assistant to create an stomata detector for plant images, see Section~\ref{sec:case-study}. The paper ends with some conclusions and further work. 

\section{A Pipeline to create deep learning-based object detectors}\label{sec:workflow}

\begin{figure*}[h]
\centering
\includegraphics[scale=.5]{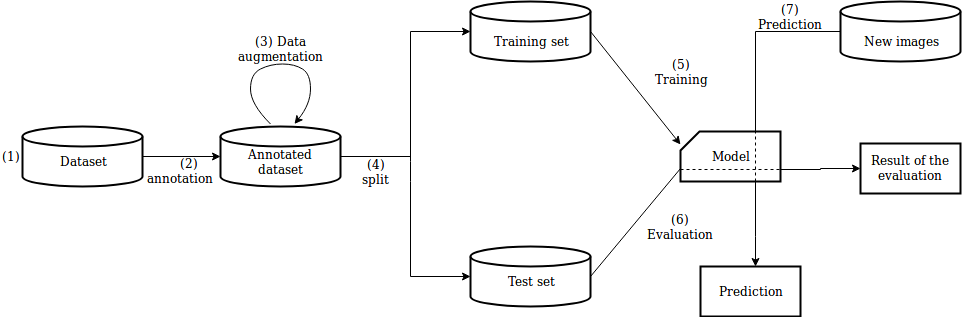}
\caption{Workflow of object detection projects}\label{fig:workflow}
\end{figure*}

In this section, we present the common workflow to create a deep learning-based object detector (see Figure~\ref{fig:workflow}), the challenges that are faced on each stage of the pipeline, the solutions that we propose to deal with those challenges, and the particularities of using the YOLO network in each stage.

\subsection*{1. Dataset acquisition}

Independently of the framework employed to create an object detection model, the first step to construct an object detection model is always the acquisition of a dataset of images containing the objects that we want to detect. This task is far from trivial since the dataset must be representative of what the model will find when deployed in the real world; and, therefore, its creation must be carefully undertaken. Moreover, acquiring data in problems related to, for instance, biomedical images might be difficult~\cite{Valle17,Asperti17}. 

There are several ways of obtaining a dataset of images depending on the project, but we can identify four main sources:

\begin{itemize}
\item \emph{Open datasets}. There are several projects that have collected a huge amount of images, such as ImageNet~\cite{ILSVRC15} or Pascal VOC~\cite{pascal-voc-2012}; however, those datasets might not contain the objects that the user is interested in detecting. 
\item \emph{Web scraping}. It is quite straightforward to create a program that uses image search engines, such as Google images or Bing images, to download images with a Creative Commons License from the web --- this allows us to legally employ those images to create object detectors. Even if this approach can be a fast way of downloading images, it might require data curation to obtain a representative dataset for the intended task.
\item \emph{Special purpose datasets}. This approach might be the only option in problems with sensitive data, like in biomedicine, where an external agent (for instant, a hospital) provides the images. The main drawbacks for this approach are the limited number of images, the difficulties to acquire new data, and restrictions related to access and use of the images.  
\item \emph{Special purpose devices to capture images}. In this case, the images are acquired by using devices that might range from a microscope to a fixed camera in a working environment. This approach has the advantage of dealing with images that are representative of the environment where the model will be later used; and, additionally, it is usually easy to obtain new images. However, the models created with those images might not be useful if the conditions change. For instance, if a model is created using images acquired with a microscope using a set of fixed conditions, the model might not work as expected if applied to images acquired with a different microscope or under different conditions. 
\end{itemize}

As we have previously mentioned, this step is independent of the tool that is employed to create the object detection model. The only restriction from the YOLO side is that the images must be stored using the JPG format. 

\subsection*{2. Dataset annotation}

Once the images have been acquired, they must be annotated, a task that is time-consuming and might require experts in the field to conduct it properly~\cite{Chest}. In the object detection context, images are annotated by providing a file with the list of bounding boxes and the category of the objects inside those boxes. Those annotation files are not written manually, but they are built using graphical tools like LabelImg~\cite{labelimg} or YOLO mark~\cite{yolo-mark}. Unfortunately, there is not a standard annotation format and, in fact, the format varies among object detection frameworks, and also among the annotation tools. Therefore, this must be taken into account when annotating a dataset of images; otherwise, a conversion step will be necessary after the annotation process is completed. 

In the case of YOLO, the annotations are provided as plain text files where each bounding box is given by the position of its center, its width and height. It is convenient to use a tool that produces the annotations directly in this format, like YOLO mark~\cite{yolo-mark}, but there are also converters from other formats; for instance, from Pascal VOC or Kitti~\cite{Weill17}.    

\subsection*{3. Dataset augmentation} As we have previously mentioned, acquiring and annotating datasets of images for object detection problems might be a challenge; this might lead to generalization problems if enough images are not acquired.  A successful method that has been applied to deal with the issue of limited amount of data is \emph{data augmentation}~\cite{Simard03}. This technique consists in generating new training samples from the original dataset by applying image transformations (for instance, applying flips, rotations, filters or adding noise). This approach has been applied in image classification problems, and there are several libraries implementing this method (for instance, Augmentor~\cite{Bloice17} or Imgaug~\cite{Jung17}).

The application of data augmentation is not straightforward in the case of object detection due to the fact that, on the contrary to data augmentation in image classification, transformation techniques alter the annotation in the context of object detection. For instance, applying the vertical flip operation to a cat image produces a new cat image --- i.e. the class of the image remains unchanged --- but the position of the cat in the new image has changed from the original image. Therefore, we have to transform not only the image but also the annotation. 

In order to apply data augmentation in object detection, the usual approach has consisted in implementing special purpose methods, depending on the 
particular problem, or manually annotating artificially generated images. Neither of these two solutions is feasible when dealing with hundreds or thousands of images. To deal with this issue, we can employ the CLoDSA library~\cite{CLODSA}, a tool that can be applied to automatically augment a dataset of images devoted to classification, localization, detection or semantic segmentation using a great variety of the classical image augmentation transformations. Using this tool, it is possible to automatically generate a considerable amount of images, together with their annotations, starting from a small dataset of annotated images. CLoDSA is compatible with the YOLO format. 

\subsection*{4. Dataset split} 

As in any other kind of machine learning project, it is instrumental to split the dataset obtained in the previous steps into two independent sets: a training set --- that will be employed to train the object detector ---  and a test set --- that will be employed to evaluate the model. Common split sizes for training and testing set include 66.6$\%$/33.3$\%$, 75$\%$/25$\%$, and 90$\%$/10$\%$, respectively. 

In the case of YOLO, the dataset split is achieved by using a particular folder structure to store the images and the labels that will be employed for training and testing, and providing two files that indicate, respectively, the set of images that will be employed for training and testing. It is worth mentioning that the YOLO framework is really sensitive to such a structure, and small changes will prevent the user from training the model. 

\subsection*{5. Training the model}

Given the training set of images, several tasks remain before starting the process to train an object-detection model. In particular, it is necessary to define the architecture of the model (that is, the number and kind of layers), fix some hyper-parameters (for example, the batch size, the number of epochs, or the momentum) and decide whether the training process starts from scratch or some pre-trained weights are employed --- the latter option, known as fine-tuning, usually improves the training process~\cite{NIPS2014_5347}.

The YOLO framework supplies several pre-defined models for training an object detector --- the best model, both in terms of accuracy and time efficiency, is the YOLO model v3. The architecture and the hyper-parameters of those models are defined in a configuration file; and, even if the by-default YOLO hyper-parameters usually work properly for training a new model, it is necessary to modify the configuration files since the model architecture must be adapted depending on the number of classes included in the dataset. Once the configuration files have been adapted, the training process can start just by executing a command. The process will end either after the provided number of epochs is reached or when the user decides to manually stop the process. In order to train a YOLO model, it is recommended to use some pre-trained weights that must be downloaded and included in the project~\cite{yolov3}. 

\subsection*{6. Evaluating the model}

After training the model, we need to evaluate it to assess its performance. Namely, for each of the images in the testing set, we present it to the model and ask it to detect the objects in the image. Those detections are compared to the ground-truth provided by the annotations of the testing set, and the result of the comparison is evaluated using metrics such as the IoU, the mAP, the precision, the recall or the F1-score~\cite{pascal-voc-2012}. The YOLO framework can perform such an evaluation automatically provided that several configuration files are correctly defined, and the correct instruction is invoked.   

A problem that might arise during the evaluation is the overfitting of the object detection model; that is, the model can detect objects on images from the training dataset, but it cannot detect objects on any others images. In order to avoid this problem, \emph{early stopping}~\cite{Sarle95} can be applied by comparing the results obtained by the models after different numbers of epochs. In the case of YOLO models, early stopping can be applied since the framework saves the state of the model every time that a certain number of training iterations is reached. 

\subsection*{7. Deploying the model}

Finally, the model is ready to be employed in images that neither belong to the training set nor to the testing set. Even if using the model with new images is usually as simple as invoking a command with the path of the image (and, probably, some additional parameters), it is worth mentioning that it is unlikely that the person who created the object detection model is also the final user of such a model. Therefore, it is important to create simple and intuitive interfaces that might be employed by different kinds of users; otherwise, they will not be able to take advantage of the object detection model.

\section{A suite of Jupyter notebooks}\label{sec:jupyter}

As we have explained in the previous section, training and using a YOLO model involves the creation and modification of several configuration files, the use of a concrete folder structure, and the execution of several commands; hence, the process is time-consuming and error prone. In this section, we introduce a simple-to-use tool that facilitates the creation of YOLO-based object detectors using Jupyter notebooks --- the suite of notebooks can be downloaded from \url{https://github.com/ancasag/YOLONotebooks}.

Jupyter notebooks~\cite{jupyter} are documents for publishing code, results and explanations in a form that is both readable and executable. Jupyter notebooks have been widely adopted across multiple  disciplines, both for its usefulness in keeping a record of data analyses, and also for allowing reproducibility. In addition, Jupyter notebooks are a useful tool to teach and learn concepts from data science and artificial intelligence~\cite{BRUNNER20161947,Ohara2015}. In our case, we have developed a suite of notebooks that guides the user in the process to create a YOLO-based object detector. The only requirement to run the notebooks is the installation of the programming language Python and its Jupyter library.

The suite of notebooks is open-source and contains several notebooks that introduce several traditional object detection notions, explain how to install the YOLO framework, and provide several examples showing how to use it (for instance, showing how the pre-trained models included in the framework might be used to detect objects in images and videos). However, the most important notebook of the suite is the one that automates the process of creating a new object detection model. Using this notebook, the user only has to (1) create a folder containing the images and the annotations in the YOLO format, and (2) fix 4 parameters in the notebook (the name of the project, the path to the folder containing the images and annotations, the list of classes, and the percentage of the images that will be employed for training); the rest of the process is carried out by simply following the steps included in the notebook. In particular, the notebook is in charge of:

\begin{itemize}
\item Validating that the images of the dataset are given in the correct format.
\item Checking that all the images of the dataset have their corresponding annotation. 
\item Guiding the user in the process to augment the dataset of images using the CLoDSA library. 
\item Splitting the dataset into a training set and testing set, and organizing those sets in the way required by the YOLO framework. 
\item Creating all the configuration files for training, evaluating and deploying the model using the last version of the YOLO network. 
\item Generating all the instructions that are required to train, test and deploy the YOLO model. 
\item Providing a simple way of invoking the generated YOLO model to detect objects in new images. 
\end{itemize}

Without our tool, all those steps should be carried out manually; hence, the burden of creating a YOLO-based object detector is significantly reduced. The aforementioned functionality has been implemented in Python using several third-party libraries such as Scikit-learn~\cite{sklearn} and OpenCV~\cite{OpenCV}.

\section{Case study: stomata detection}\label{sec:case-study}

In this section, we show the feasibility of using our tool by creating a stomata detector in images of plant leaves. A stoma is a tiny opening, or pore, that is used for gas exchange in plants. The amount and behavior of stomata provide key information about water stress levels, production ratio, and, in general, the overall health of the plant~\cite{Buttery93}. Hence, by measuring the number of stomata, it is possible to manage better the resources in agriculture and obtain better yields~\cite{Giorio99}. However, manually counting the number of stomata is a time-consuming and subjective task due to the considerable amount of stomata in an image, their small size, and the fact that it is necessary to analyze batches of dozens of images. Therefore, it is useful to automatically detect and count the number of stomata in plant images.  

Several studies can be found on automatic detection of stomata, but they employ traditional features like Haar~\cite{HaarStomata}, MSER~\cite{MSERStomata} or HOG~\cite{HOGStomata} combined with a cascade classifier, and neither the implementation of those techniques nor the datasets of those studies are available, making unfeasible the reproducibility of their results  or the use of their models with new images. On the contrary, using the suite of notebooks presented in the previous section, and a dataset of images provided by the University of Missouri, we have built a freely available YOLO-based stomata detector for images of plant leaves.  

The dataset employed to construct our model consists of 468 microscope images of the leaf epidermises of different plant types (including cotton, peanut and maize plants), and those images contain a total amount of 1652 stomata --- the dataset is available from the authors on request. The images were annotated by expert biologists using the LabelImg program, that produces annotations in the Pascal VOC format; hence, it was necessary to transform those annotation to the YOLO format using a Python script. After storing the images and labels in the same folder, and fixing the 4 parameters explained in the previous section, the following process was conducted guided by the notebook.   

First of all, to improve the generalization of the model, the CLoDSA library was employed to augment the dataset of images by employing techniques like flips, rotations, filters and adding noise. The final dataset was formed by a total of 4212 images, enough for training an object detector. Subsequently, the dataset was split into two sets, using 90$\%$ of the dataset for training, and $10\%$ for testing --- those sets were automatically split, organized in the corresponding folders, and, additionally, all the configuration files were automatically generated. After training the model for 250000 epochs, we stopped the process and evaluated the model using the testing set. The results achieved by the model were a mAP of $90.91\%$, a precision of $98\%$ and a F1-score of $99\%$. 

\begin{figure}
\centering
\includegraphics[scale=0.15]{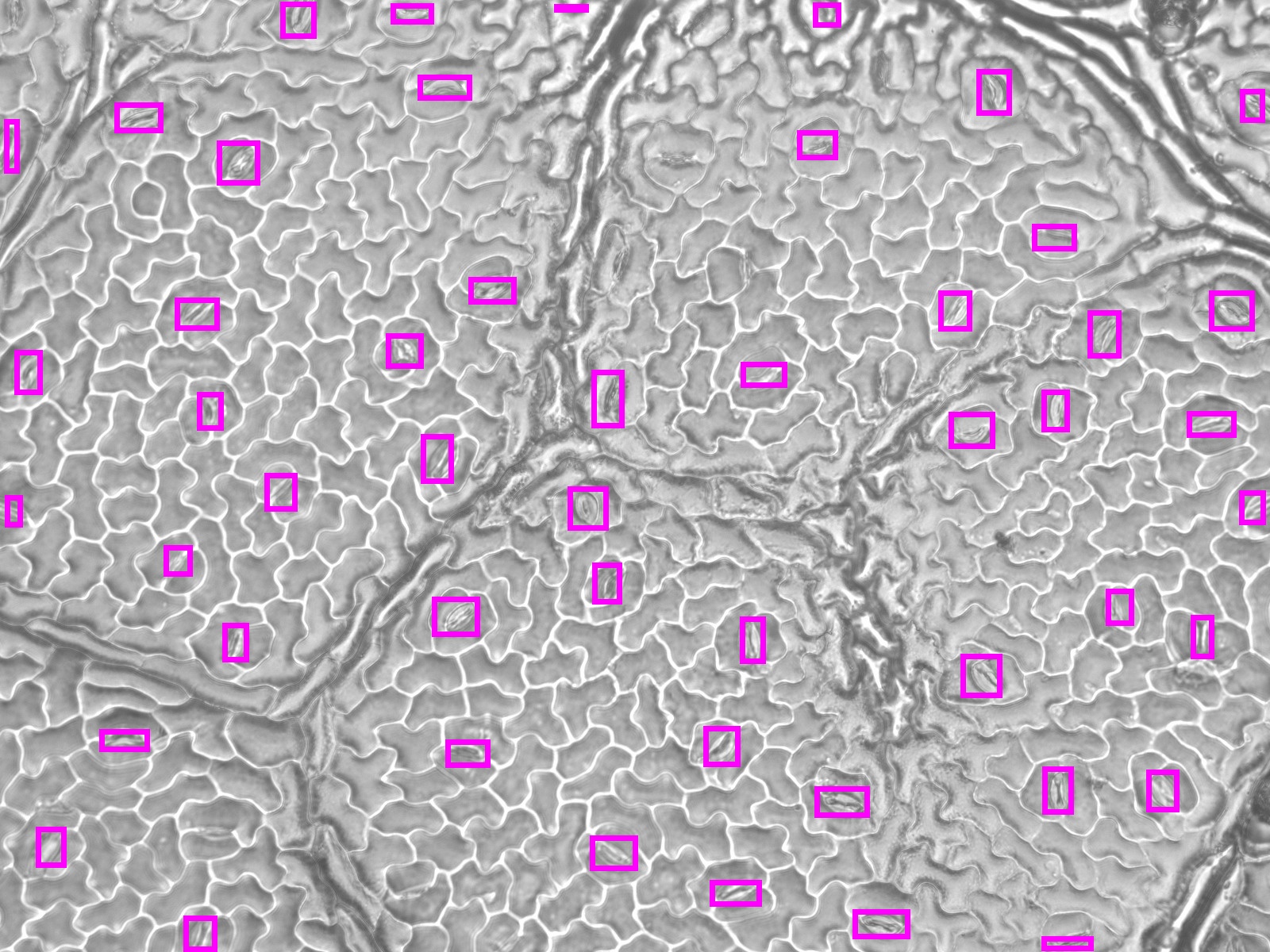}
\caption{Stomata identification results}\label{fig:prediction}
\end{figure}

This process can be easily reproduced using the notebook available in the project webpage, and the obtained model can be invoked to detect stomata in images that do not belong neither to the training nor the testing set, see Figure~\ref{fig:prediction}. The model was trained using a Titan Xp GPU, but it can be employed for detecting stomata in any computer that has Python and C++ installed on it. 

\section{Conclusions and further work}

Democratizing Artificial Intelligence is a movement that has been born with the goal of making artificial intelligence accessible to non-expert users from different fields~\cite{PANDA,AutoML}. The work presented in this paper can be framed in that context; in particular, after carefully analyzing all the steps that are required to construct an object detector using deep learning techniques, we have presented a suite of Jupyter notebooks that allows non-expert users to easily create fast and accurate object-detection models using the YOLO approach --- one of the most effective methods for object detection. The suitability of our approach has been tested by developing a model for stomata detection in plant images that achieves a mAP of $90.91\%$. 

In addition to the benefit of simplifying the creation of object detection models, the suite of Jupyter notebooks has value as a teaching material since we have included explanations, both textual and graphical, of all the steps involved in the creation of an object detector; and, hence, they can be useful for Artificial Intelligence and Computer Vision courses.

The main drawback of our suite of plugins is that users need a GPU installed, and properly configured, in their computer, since training deep learning models usually requires the use of a GPU. Therefore, as further work, we plan to integrate our tool in a cloud service like Amazon or Google Cloud to provide a cheap and fast way of constructing object detection models. Moreover, we intend to extend the suite of plugins with other deep learning techniques for object detection, such as SSD or faster R-CNN; and also for other computer vision problems like image classification or semantic segmentation.

\bibliographystyle{IEEEtran}
\bibliography{biblio}

\end{document}